\documentclass[letterpaper, 10 pt, conference]{ieeeconf}
\IEEEoverridecommandlockouts
\title{\LARGE \bf Continual Domain Randomization} %

\author{Josip Josifovski$^{1*}$, Sayantan Auddy$^{2*}$, Mohammadhossein Malmir$^{1}$,\\ Justus Piater$^{2,4}$, Alois Knoll$^{1}$ and Nicol\'as Navarro-Guerrero$^{3}$%
\thanks{$^{1}$ School of Computation Information and Technology, Technical University of Munich, Germany.}%
\thanks{$^{2}$ Department of Computer Science, University of Innsbruck, Austria.}%
\thanks{$^{3}$ L3S Research Center, Leibniz Universit\"at Hannover, Germany.}%
\thanks{$^{4}$ Digital Science Center (DiSC), University of Innsbruck, Austria.}
\thanks{$*$ Equal contribution}
\thanks{Sayantan Auddy is supported by a doctoral scholarship granted by the University of Innsbruck, Vice-Rectorate for Research.}
}

\usepackage{cite}  %
\usepackage{amsmath,amssymb,amsfonts}
\usepackage{mathtools}
\usepackage{algorithmic}
\usepackage{graphicx}
\usepackage{textcomp}
\usepackage{xcolor}
\usepackage{xfrac}
\usepackage{booktabs} %
\usepackage{url}
\usepackage{hyperref}

\usepackage{macros} %
\usepackage{glossaries}
\usepackage{paralist}
\usepackage{balance} %
\usepackage{cite}
\usepackage{wasysym}
\usepackage{tcolorbox} %
\usepackage[absolute]{textpos}

\newacronym{ROS}{ROS}{Robot Operating System}

\hypersetup{
  pdftitle={Continual Domain Randomization},%
  pdfauthor={Josifovski, Auddy, Malmir, Piater, Knoll, and Navarro-Guerrero},%
  pdfsubject={Robotics (cs.RO); Artificial Intelligence (cs.AI)},
  pdfkeywords={sim2real, Reality Gap, Reinforcement Learning, Sequential Randomization, Real-World Robotic Tasks},%
  colorlinks=true,
  linktoc=all,
  linkbordercolor=white,
  linkcolor=black,
  citecolor=black,
  urlcolor=black,
}

\begin{document}

\maketitle
\thispagestyle{empty}
\pagestyle{empty}

\begin{abstract}
Domain Randomization (DR) is commonly used for \simtoreal{} transfer of reinforcement learning (RL) policies in robotics. Most DR approaches require a simulator with a fixed set of tunable parameters from the start of the training, from which the parameters are randomized simultaneously to train a robust model for use in the real world. 
However, the combined randomization of many parameters increases the task difficulty and might result in sub-optimal policies. 
To address this problem and to provide a more flexible training process, we propose \emph{Continual Domain Randomization}~(CDR) for RL that combines domain randomization with continual learning to enable sequential training in simulation on a subset of randomization parameters at a time. Starting from a model trained in a non-randomized simulation where the task is easier to solve, the model is trained on a sequence of randomizations, and continual learning is employed to remember the effects of previous randomizations. 
Our robotic reaching and grasping tasks experiments show that the model trained in this fashion learns effectively in simulation and performs robustly on the real robot while matching or outperforming baselines that employ combined randomization or sequential randomization without continual learning.  Our code and videos are available at \codeurl{}.

\blockcomment{
\sa{Draft of alternate abstract}: 
Randomization is a commonly used method for Sim2Real transfer in robotics, and often multiple robot parameters are randomized together in an attempt to create a simulation model that is robust to the stochastic nature of the real world. However, at the onset of training in simulation, it may not be clear which parameters should be randomized. Moreover, excessive randomization can prevent the robot model from learning in simulation. In this work, we show that we can start from a model trained in ideal simulation, and then randomize one parameter at a time with continual learning to remember the effects of previous randomizations. At the end of all the randomizations, we get a model that learns effectively in simulation and also performs more robustly in the real world compared to the other baselines.
\TODO{Mention reacher, grasper. Summarize key findings. Code to be shared?}

\TODO{Abstract placeholder from old paper, change last}Randomization is currently a widely used approach in Sim2Real transfer for data-driven learning algorithms in robotics. Still, most Sim2Real studies report results
for a specific randomization technique and often on a highly
customized robotic system, making it difficult to evaluate
different randomization approaches systematically. To address
this problem, we define an easy-to-reproduce experimental
setup for a robotic reach-and-balance manipulator task, which
can serve as a benchmark for comparison. We compare four
randomization strategies with three randomized parameters
both in simulation and on a real robot. Our results show that
more randomization helps in Sim2Real transfer, yet it can also
harm the ability of the algorithm to find a good policy in
simulation. Fully randomized simulations and fine-tuning show
differentiated results and translate better to the real robot than
the other approaches tested.
}

\end{abstract}


\section{Introduction}

Deep reinforcement learning (DRL) in robotics research relies heavily on simulated environments, mainly because of the sample inefficiency of DRL algorithms and the absence of safe exploration guarantees, which make it impractical to train such algorithms on physical robots directly. While simulations provide virtually unlimited training data and eliminate safety concerns during training, their approximative nature can significantly degrade the model's performance on the real system in \simtoreal{} transfer~\cite{Zhao2020SimtoReal} -- a phenomenon known as the \emph{reality gap}.

To bridge the reality gap, one can try to attain \emph{zero-shot} \simtoreal{} transfer by crafting an accurate simulation and perfectly mimicking the behavior of the real system~\cite{Zhao2020SimtoReal}.
Unfortunately, even for elementary systems, simulating reality with sufficient precision is rarely possible. 
Alternatively, a simulation-trained DRL model can be  \emph{finetuned} on the target to improve performance. This \emph{domain adaptation} approach updates the policy learned in simulation and usually leads to improved performance. However, it makes the policy system-specific and requires at least some real-world training, which can be unsafe and time-consuming \cite{dulac2021challenges}.

Due to the impracticality of developing perfect simulations and the limitations of domain adaptation, recent works on \emph{domain randomization}~\cite{Tobin2017Domain} rely on the randomization of simulation parameters. Here, instead of accurately simulating every parameter of the real system, the relevant simulation parameters are randomized to cover the difference between the real world and simulation. 
However, it can be challenging to precisely identify all the essential randomization parameters at the start. Moreover, the excessive randomization of many parameters jointly increases the task complexity and the agent's uncertainty, making it difficult to find a good policy in simulation~\cite{josifovski2022analysis}.

To address these limitations, we propose \emph{Continual Domain Randomization} (CDR), which combines \emph{domain randomization}~\cite{Tobin2017Domain} with \emph{continual learning}~\cite{parisi2019continual} (CL) for sim2real transfer. 
Viewing domain randomization as a CL problem provides greater flexibility in the initial choice of the randomization parameters. Additionally, randomizing a subset of all parameters at a time reduces the risks of excessive randomization.
In CDR, we start with an agent trained in an idealized simulation (with all randomizations disabled) and successively train the agent in differently randomized simulations. We treat each randomization as a separate CL \emph{task} and enable the DRL agent to continually train under each randomization without \emph{catastrophically forgetting}~\cite{parisi2019continual} the knowledge gained from previous training runs (i.e., past \emph{tasks}). We implement an offline and an online version of CDR by combining the Proximal Policy Optimization (PPO)~\cite{Schulman2017Proximal} algorithm with the regularization-based CL algorithm \emph{Elastic Weight Consolidation} (EWC)~\cite{kirkpatrick_overcoming_2017} and its online variant~\cite{Schwarz2018Progress}. Nevertheless, CDR is a general strategy that can be implemented with other DRL or CL algorithms.

\begin{textblock}{13.24}(1.7,0.3)
\makebox[0cm][l]{
\begin{minipage}{\dimexpr0.9\textwidth}
    \begin{tcolorbox}[colframe=black, colback=white, sharp corners=all, boxrule=0.15mm, width=\textwidth, boxsep=0mm]
        \footnotesize \copyright{} 2024 IEEE. Personal use of this material is permitted.  Permission from IEEE must be obtained for all other uses, in any current or future media, including reprinting/republishing this material for advertising or promotional purposes, creating new collective works, for resale or redistribution to servers or lists, or reuse of any copyrighted component of this work in other works.
    \end{tcolorbox}
\end{minipage}}
\end{textblock}

We perform experiments for \simtoreal{} transfer of two different robotics tasks, reaching and grasping, and compare the real-world performance of policies learned using CDR against \textit{finetuning}, \textit{full-randomization}, and the ideal model trained without randomization. 
Our results show that CDR can match or outperform baselines that jointly randomize all parameters throughout the training process. 
At the same time, it is more stable and less sensitive to the order of randomizations than sequential baselines like \textit{finetuning}, which only optimizes the policy for the newest randomization and ignores previous ones.
These results highlight the potential of CDR as a general and flexible framework for helping bridge the reality gap in DRL-based robotics.

\section{Related Work}

Even though there are some cases of \simtoreal{} transfer where measuring and precisely simulating all relevant parameters of the real system may be possible~\cite{James20163D, Kaspar2020Sim2Real}, 
due to wear and tear in the robot's hardware or changes in external conditions like temperature or humidity, they can vary significantly~\cite{Zhao2020SimtoReal}. 
Intuitively, providing some degree of variability in the simulator parameters and training a model that can deal with a range of parameter values would lead to a more robust \simtoreal{} transfer~\cite{Muratore2022Robot}. 
For this reason, \emph{domain randomization} (DR)~\cite{Tobin2017Domain, Josifovski2018Object, Peng2018SimtoReal, OpenAI2019Solving, Mehta2020active, ramos2019bayessim, muratore2022neural, huang2023what} has become a widely used \simtoreal{} technique. %
However, inappropriate and excessive randomization can increase the uncertainty of the system and the difficulty of solving the task in simulation, leading to sub-optimal policies and longer training times~\cite{josifovski2022analysis}.

One approach to circumvent those problems is \emph{automatic domain randomization} \cite{OpenAI2019Solving}. Here, 
the randomization increases progressively as the policy under the current randomization improves the performance. 
In contrast, \emph{active domain randomization}\cite{Mehta2020active} starts from wide randomization ranges and learns the most informative ranges for training a robust policy. While the previous approaches assume that only simulated data is available, data from the real system can be used in offline fashion~\cite{tiboni2023dropo, muratore2022neural} to find suitable parameter distributions, or to identify which parameters are relevant for \simtoreal{} transfer through causal discovery~\cite{huang2023what}.

Continual learning (CL) algorithms enable knowledge accumulation from a sequence of tasks without \emph{catastrophic forgetting}~\cite{parisi2019continual}.
CL has traditionally targeted supervised learning scenarios using static datasets or reinforcement learning problems with discrete actions~\cite{kirkpatrick_overcoming_2017, Schwarz2018Progress}. 
More recently, CL has been applied to more challenging scenarios such as robot learning, both in 
simulation~\cite{schopf_hypernetwork-ppo_2022}
as well as in the real world~\cite{auddy2023continual, auddy2023scalable}. Continual reinforcement learning (CRL) in continuous action and observation spaces~\cite{khetarpal_towards_2020} has also emerged as a popular means of training robots to learn multiple tasks. 
CRL scenarios can be of different types~\cite{khetarpal_towards_2020}:
\begin{inparaenum}[(i)]
\item the state transition and action space for different tasks remain the same, but the reward function changes (the agent needs to perform fundamentally different actions such as pushing, reaching, etc., for each task)~\cite{luo2023relay};
\item the state transition function changes, but the reward function and action space remain the same. Here, the environment dynamics change, but the ultimate goal of the agent remains the same~\cite{Josifovski2020Continual}.
The \simtoreal{} scenario belongs to the latter. 

\emph{Progressive networks}~\cite{rusu2016progressive} have also been used for continual \simtoreal{} transfer~\cite{Rusu2017SimtoReal}. For instance, as shown by Rusu et al.,~\cite{Rusu2017SimtoReal}, an initial neural network can be trained in simulation to solve a task, and a new network, with lateral connections to the previous network for feature reusing, is instantiated and trained on the real robot.  
Another approach used for \simtoreal{} transfer is \textit{Policy distillation}~\cite{rusu2015policy}. For instance, Traor{\'e} et al.,~\cite{traore2019continual} trained a separate policy and generated a dataset of observation-action pairs for each task in simulation. The datasets are then used in a supervised learning fashion to train a single model that can solve the tasks in the real environment.

Unlike other domain randomization approaches, in CDR no fixed set of randomization parameters during training is assumed -- the model can train on a subset of randomization parameters at a time and adapt to new randomization parameters later. This is achieved by combining sequential randomization with continual learning, which retains the effects of previous randomizations and avoids overfitting to the last set of parameters. Unlike other continual sim2real approaches, CDR uses a single neural network trained across all tasks. Additionally, the online CDR variant avoids the linear increase in memory for keeping a separate dataset or network parameters per task. Finally, CDR can easily be combined with methods like \emph{automated} \cite{OpenAI2019Solving} or \emph{active} \cite{Mehta2020active} DR to find suitable ranges for the subset of randomization parameters within a single training run in the sequence.

\end{inparaenum} 

\section{Methodology}\label{sec:method}

\begin{figure*}
    \vspace{4px}
    \includegraphics[width=\textwidth]{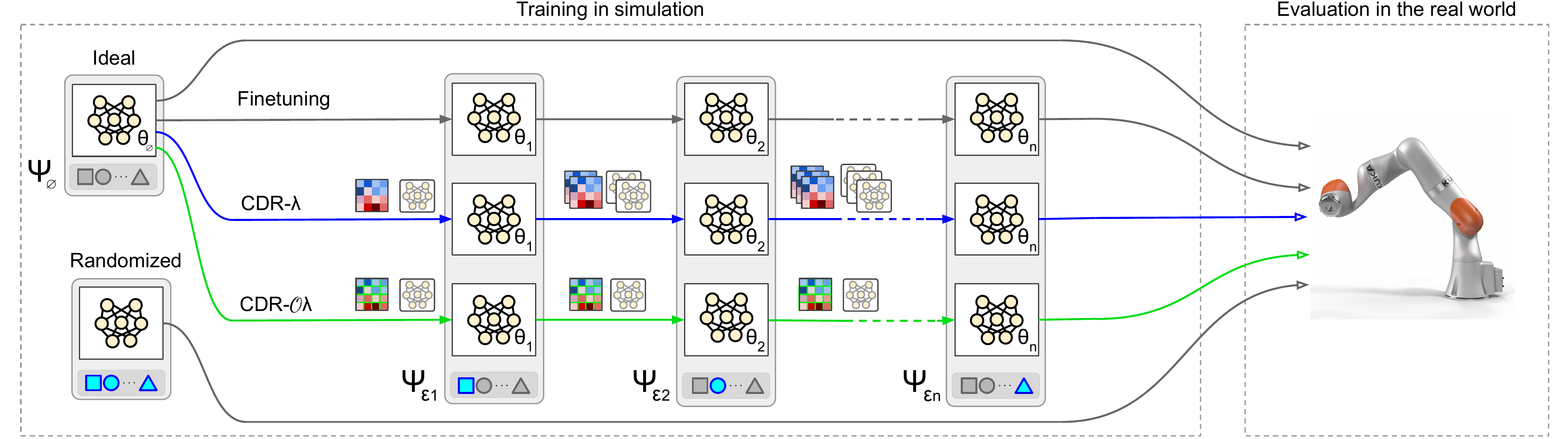}
    \caption{Overview of our proposed CDR approach. \cdrewc{} (\textcolor{blue}{blue} arrow) uses a set of network snapshots and Fisher matrices (one for each past task) for continual learning, while \cdronewc{} (\textcolor{green}{green} arrow) uses only a single parameter snapshot and Fisher matrix. Other baselines are shown with \textcolor{gray}{gray} arrows.
    Each shape ($\square,\Circle, \cdots,  \triangle$) represents a unique randomization parameter set. Disabled and enabled sets are indicated with gray and blue respectively.}
    \label{fig:overview}
        \vspace{-4px}
\end{figure*}

\subsection{Problem Description}

Let $\mathcal{M}=(\mathcal{S}, \mathcal{A}, P, R, S_0, \gamma, T)$ define a Markov Decision Process (MDP) where $\mathcal{S}$ denotes the state space, $\mathcal{A}$ is the action space, $P(s_{t+1}\in\mathcal{S}|s_{t}\in\mathcal{S}, a_t \in \mathcal{A})$ is the state transition probability, $R$ is the reward function, $S_0$ defines an initial state distribution, $\gamma$ is the reward discount factor and $T$ defines the finite horizon. The objective of reinforcement learning is to find the parameters $\theta$ for a parameterized policy $\pi_\theta(a_t|s_t)$ to maximize the expected return $\mathbb{E}_{\pi_\theta}[ \sum^{T}_{t=0} \gamma^t R(s_t, a_t)]$ where $s_0 \sim S_0$, $s_{t+1} \sim P$ and $a_t \sim \pi_\theta(a_t|s_t)$.
Let $\Phi$ denote a real system with unknown dynamics, and $\mathcal{E} = \{\epsilon_1, \cdots, \epsilon_n\}$ denote the set of all randomization parameters of a simulator $\Sigma$ of $\Phi$. The goal is to optimize $\theta$ using $\Sigma$ such that it maximizes the expected return on $\Phi$ (zero-shot sim2real transfer).

\subsection{Continual Domain Randomization}

In our suggested approach, \emph{Continual Domain Randomization}~(CDR), the policy parameters $\theta=\tilde{\theta}$ are initialized and optimized in an idealized simulation $\Psi_\varnothing$ (with all randomizations disabled). Afterward, we sequentially optimize the resulting $\theta$ with CL on a sequence of simulations where a subset of randomization parameters is enabled each time, as shown in \myfigure{fig:overview}:
$\tilde{\theta} \xrightarrow[]{\Psi_\varnothing} \theta_\varnothing \xrightarrow[]{\Psi_{\epsilon_1}} \theta_1 \cdots  \xrightarrow[]{\Psi_{\epsilon_n}} \theta_n$.
\medskip

\noindent\textbf{\cdrewc{}}: combines the PPO~\cite{Schulman2017Proximal} algorithm with the popular regularization-based CL algorithm Elastic Weight Consolidation (EWC)~\cite{kirkpatrick_overcoming_2017}, $\lambda$ denotes the regularization process with EWC. 
EWC is an approximate Bayesian method that quantifies the importance of neural network parameters for remembering previous knowledge using the empirical \emph{Fisher information matrix} (FIM).
The FIM is used to estimate the posterior distribution over the network parameters, approximating it as a multivariate Gaussian centered around the maximum a posteriori (MAP) estimate. The diagonal of the covariance matrix is then used to determine parameter importance.
The loss function for learning task $C$ after tasks $A$ and $B$ is:
\begin{equation}
    \mathcal{L}(\theta) = \mathcal{L}_C(\theta) + 
     \cfrac{\lambda_B}{2}  ||\theta -\theta^*_{B}||^2_{F_{B}} +
     \cfrac{\lambda_A}{2}  ||\theta -\theta^*_{A}||^2_{F_{A}}
     \label{eq:ewc_loss}
\end{equation}
where $\mathcal{L}_C$ is the task-specific loss for task $C$, $\lambda_A$ and $\lambda_B$ are the regularization constants, $\theta^*_A$ and $\theta^*_B$ are the saved MAP parameters after learning tasks $A$ and task $B$ respectively, and $F_A$ and $F_B$ are the diagonal Fisher information matrices for the first two tasks. 
In this paper, we adapt PPO for CL by regularizing the parameters of \emph{only} the actor network with EWC.
After learning each task, we roll out trajectories of the PPO agent trained on the most recent environment and store these in a replay buffer. After that, we compute the (diagonal) empirical Fisher information for the latest task using the squared gradients of the log-likelihoods of the action~\cite{kessler2020unclear} predicted by the PPO policy: 
\begin{equation}
    F(\theta) = \cfrac{1}{n} \sum\limits_i^n \nabla_\theta \log p_\theta(y_i|x_i) \nabla_\theta \log p_\theta(y_i|x_i)^\mathrm{T}
    \label{eq:ewc_fisher}
\end{equation}
\noindent here, $n$ denotes the number of samples in the replay buffer, $x_i$ and $y_i$ denote the input and output of the policy network, and $\theta$ denotes the parameters of the policy network. Since the critic does not contribute to the log-likelihood computation, the parameters of the critic are not regularized. In contrast to previous implementations of PPO and EWC~\cite{gaurav2020safety}, but similar to online-EWC~\cite{Schwarz2018Progress}, we normalize the elements of $F$ to be in the range [0.0,1.0] after each task. This normalization decouples the regularization constant $\lambda$ from the arbitrary scaling of the Fisher information of individual tasks and allows us to use of a single value of $\lambda$ for all tasks.

A disadvantage of regular EWC is that for each randomization (task), a separate snapshot of the actor network parameters and the corresponding empirical diagonal Fisher information needs to be saved. 
This results in the undesired linear scaling of storage requirements. To counter this, we suggest combining PPO with online-EWC~\cite{Schwarz2018Progress} creating \textbf{\cdronewc{}}.
After learning a task, a replay buffer is filled, and the empirical Fisher information is computed with Eq.~\ref{eq:ewc_fisher}. Thus, instead of maintaining a separate copy of the network and Fisher information for each task, only one copy of the network after the most recent task is kept, and the network is updated using
\begin{equation}
    \mathcal{L}_i(\theta)=\tilde{\mathcal{L}}_i(\theta) + \cfrac{\lambda}{2}||\theta - \theta_{i-1}^*||^2_{\gamma F^*_{i-1}}
\end{equation}
\noindent where $\mathcal{L}_i(\theta)$ is the overall loss for the current task, $\tilde{\mathcal{L}}_i(\theta)$ is the task-specific PPO loss, $\theta_{i-1}^*$ is the snapshot of the network after the previous task, and $F^*_{i-1}$ is the \emph{online} Fisher information after the previous task. $F^*_{i-1}$ and the Fisher information of the current task $F_i$ are used to update the online Fisher information $F_i^*$ using a hyperparameter $\gamma$:
\begin{equation}
    F_i^* = \gamma F^*_{i-1} + F_i
\end{equation}
\noindent We again normalize $F_i$ to be in the range [0.0,1.0] to avoid any arbitrary scaling. Similar to \cdrewc{}, we only regularize the PPO actor network for \cdronewc{}.

\subsection{Baselines}

We implement multiple baselines~\cite{josifovski2022analysis} with PPO~\cite{Schulman2017Proximal} (\myfigure{fig:overview}):
\begin{enumerate}[(i)]
    \item \textbf{Ideal}: Training is conducted in the \emph{ideal} simulation environment without any randomization.
    \item \textbf{Finetuning}: The model trained in the ideal simulation is used as a starting point and then successively \emph{finetuned} on each of the randomization parameters, one at a time, without any continual learning.
    \item \textbf{Randomized}: The model is trained by activating all the randomization parameters together at the same time.
\end{enumerate}

\section{Experiment Setup}

For the experiments, we consider the \emph{reach-and-balance}~\cite{josifovski2022analysis} task with a 2-joint-controlled robotic arm, and a robotic \emph{grasping} task with a kinematically redundant, 7-joint-controlled arm and a three-fingered gripper~\cite{gripper}. Both tasks are defined in continuous state and action spaces, and the RL agent exercises fine-grained direct control over the robot's joint velocities.
We intentionally use joint space control, 
because it requires minimal information about the real system's dynamics \cite{varin2019comparison}. Moreover, in this case the agents are supposed to learn to compensate for dynamic effects, e.g., inertial forces  \cite{varin2019comparison} through randomization, which is particularly important and part of our \simtoreal{} transfer analysis.

\subsection{Reacher Task}
The \emph{reacher} task is defined as:
\begin{equation}
\mathbf{s}_t = [\Delta{x}_t, \Delta{y}_t, \Delta{z}_t,{q}_1,{q}_2] \in \mathbb{R}^5
\end{equation}
\begin{equation}
\mathbf{a}_t = [\dot {q}_1^{\prime},\dot{q}_2^{\prime}] \in \mathbb{R}^2
\end{equation}
\begin{equation}
r_t = 
    \begin{cases}
    - d_t   &\text{ if $C$ is false} \\
    - d_t \times (T - t) &\text{ otherwise}
    \end{cases}
    \in \mathbb{R}
\end{equation}
\noindent where $\mathbf{s}_t$ is the agent's state at timestep $t$, $\Delta{x}$, $\Delta{y}$ and $\Delta{z}$ are the distances between the end-effector and the target along the corresponding coordinate axes, and ${q_i}$ are the joint positions. $\mathbf{a}_t$ is the set of joint velocities the agent commands at timestep $t$. The reward $r_t$ at each timestep is defined as the negative Euclidean distance ${d}_t$ between the end-effector and the target, unless the robot collides with the floor or reaches joint limits, in which case the terminal condition $C$ becomes true, the episode terminates and the reward is the negative distance at that timestep multiplied with the remaining timesteps to the maximal episode length $T$.

\subsection{Grasper Task}
The \emph{grasping} task is defined as:
\begin{equation}
\begin{aligned}
\mathbf{s}_t = [ & \Delta{x}_t, \Delta{y}_t, \Delta{z}_t, {q}_1,{q}_2,{q}_3,{q}_4,{q}_5,{q}_6,{q}_7,  \\
& \Delta{fx}_t, \Delta{fy}_t, \Delta{fz}_t,  \Delta{ux}_t, \Delta{uy}_t, \Delta{uz}_t] \in \mathbb{R}^{16}
\end{aligned}
\end{equation}
\begin{equation}
\mathbf{a}_t = [
\dot {q}_1^{\prime},\dot{q}_2^{\prime},
\dot {q}_3^{\prime},\dot{q}_4^{\prime},
\dot {q}_5^{\prime},\dot{q}_6^{\prime},
\dot {q}_7^{\prime}
] \in \mathbb{R}^7
\end{equation}
\begin{equation}
r_t = 
    \begin{cases}
    - {dp}_t - 0.5 \times {du}_t - c    & \text{ if $t<T$} \\
    - {dp}_t - 0.5 \times {du}_t - c + g & \text{if $t=T$} \\
    \end{cases}
    \in \mathbb{R}
\end{equation} 
\noindent where $\mathbf{s}_t$ is the state of the agent at timestep $t$, $\Delta{x_p}$, $\Delta{y_p}$ and $\Delta{z_p}$ are the distances between the end-effector and the center of the object along the corresponding coordinate axes, and ${q_{1-7}}$ are the joint positions. 
$\Delta{fx}$, $\Delta{fy}$, $\Delta{fz}$ are the distances between the \textit{forward} unit direction vector of the object and the \textit{forward} unit direction vector of the end-effector along each axis, and $\Delta{ux}$, $\Delta{uy}$, $\Delta{uz}$ are the distances between the \textit{up} unit direction vector of the plane on which the object is placed and the \textit{up} unit direction vector of the end-effector along each axis. 
$\mathbf{a}_t$ is the set of joint velocities the agent commands at time $t$. 
Each episode has a fixed duration $T$. If a collision occurs or the robot reaches joint limits during the episode, zero velocities are commanded until the episode ends. 
At the end of the episode, if there were no collisions and the joint limits were not reached, the robot stops moving and the gripper closes in an attempt to grasp the object. 
The reward at timestep $t$ is defined in terms of the negative Euclidean distance ${dp}_t$ between the end-effector and the center of the object, the negative Euclidean distance ${du}_t$ between the \textit{up} unit vectors of the gripper and the floor, and a penalty coefficient $c$ that is 0 by default and 1 if there is a collision or joint limits are reached during the episode. At the end of the episode, if the grasping attempt was successful, the term $g$ is set to 10, otherwise it is 0.

\subsection{Robot Platform}
\begin{figure}[t]
 \vspace{5px}
    \centering
	\includegraphics[width=0.48\textwidth]{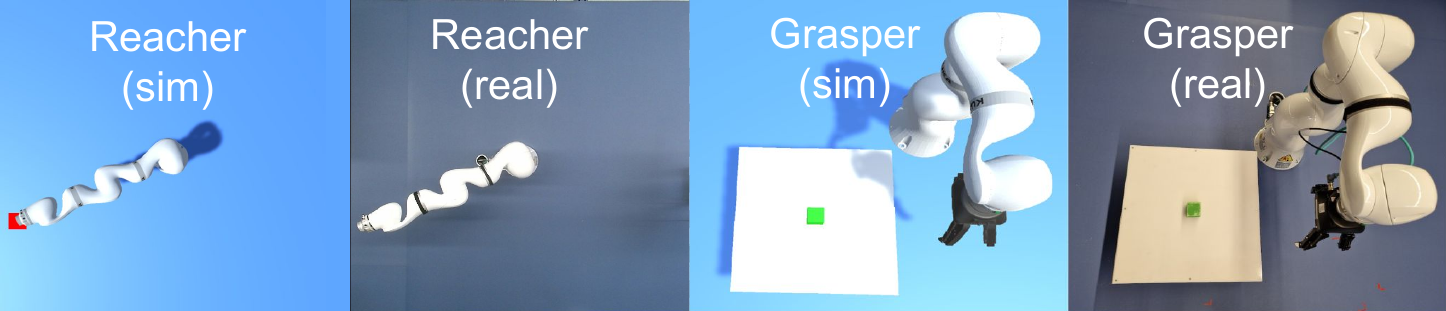}
	\caption{The simulated and real environments for reaching and grasping.}
    \label{fig:sim_real_envs}
\end{figure}

The simulated and real-world environments are shown in \myfigure{fig:sim_real_envs}. We use the Unity-based platform
introduced in~\cite{Josifovski2020Continual} for the simulation. For both \emph{reacher} and \emph{grasper}, we use the \emph{KUKA LBR iiwa 14} robotic manipulator~\cite{sturz2017parameter}, and for the grasping task we use the 3-Finger Adaptive Robotic Gripper (3F model) by Robotiq %
in pinch mode, relying on the gripper parameters and meshes from the URDF data provided by the ROS-Industrial~\cite{edwards2012ros} package. The real robot is controlled with the IIWA stack \cite{Hennersperger2017MRIBased} via \gls{ROS} \cite{ros-operating-system}. The box used for the grasping experiments is a cube with a side of 50mm. 

\begin{figure}[b]
    \centering
	\includegraphics[width=\columnwidth]{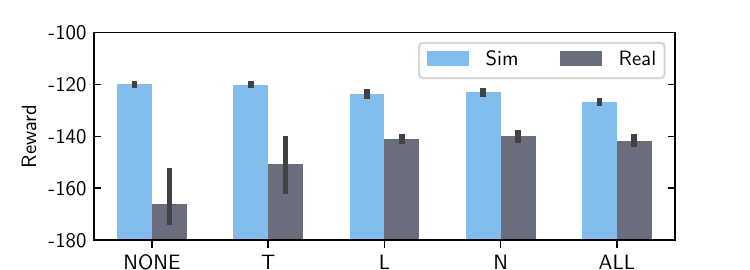}
	\caption{Effects of different randomization parameters on \simtoreal{} transfer for the reaching task.}
    \label{fig:parameter_importance}
\end{figure}

\subsection{Training Procedure}
\label{training_procedure}
We use three randomisation parameters commonly used in the literature~\cite{josifovski2022analysis}, \cite{Peng2018SimtoReal}: 
\begin{inparaenum}[(i)]
    \item Latency $L$: random delay in observing the current state;
    \item Torque $T$: the joint stiffness and damping coefficients are randomized, resulting in a randomized torque for reaching the commanded velocity; and 
    \item Noise $N$: noise drawn from a uniform distribution is randomly incorporated to simulate sensor inaccuracies.
\end{inparaenum}
The implementation is identical to~\cite{josifovski2022analysis} and the ranges of the randomization parameters are provided in \mytable{table:hyperparameters_details}.

The efficacy of \simtoreal{} transfer under sequential training strategies like finetuning or CDR may be influenced by the ordering of randomization parameters, as each parameter can have varying effects on the transfer process~\cite{josifovski2022analysis}.
Instead of unnecessarily testing all possible permutations of $L$, $N$, and $T$, we empirically evaluate the importance of each parameter for \simtoreal{} transfer on the reaching task
to determine the best and worst case ordering scenario. In the best scenario, the parameter that leads to the best \simtoreal{} transfer is at the end of the training sequence, while in the worst case scenario, it is at the beginning. For the empirical evaluation, we perform training on a single-parameter randomized simulation from scratch and evaluate the models in ideal simulation and on the real system. The results from 5 independent seeds for each parameter, as well as the baselines where no parameter or all parameters are randomized, are presented in \myfigure{fig:parameter_importance}. 
Based on the results, we define 
$TLN$
and 
$NLT$ orderings for the sequential training strategies in the experiments.

\blockcomment{
\begin{table}[b]
    \begin{center}
    \caption{Training hyperparameters.}
    \label{table:hyperparameters_details}
    \begin{tabular}{p{20mm} llll}
    \hline\noalign{\smallskip}
    Param. & Value & Param. & Value \\
    \hline\noalign{\smallskip}
     \specialcell{Bias-variance \\trade-off ($\lambda$)}  &  0.95 &  Torque($T$) & \specialcell{stiffness: [$10^1$, $10^3$]\\damping: [$10^1$, $10^3$]}\\
     Discount factor ($\gamma$)   &  0.99 & Latency($L$) & \specialcell{[0, 1] sec}\\ 
     Learning rate ($\alpha$)  & 0.00025 & Noise($N$) & \specialcell{[0, 10] \%}\\
     \specialcell{Value function \\coefficient ($c_{1}$)}  &  0.5 & EWC-$\lambda$ & $5*10^3$\\
     Entropy coefficient ($c_{2}$) & 0.0 & \specialcell{EWC buffer\\ limit} & 100\\
     Clip range ($\epsilon$)  &  0.1 & \specialcell{EWC reply \\batch size} & 32\\ 
     \specialcell{Optimization epochs}  &  10 & EWC online $\gamma$ & 0.95 \\ 
     Mini-batches  &  32 & \specialcell{EWC reply \\samples} & 5000\\ 
    \hline\noalign{\smallskip}
    \end{tabular}
    \end{center}
\end{table}
}

\begin{table}[t]
\vspace{4px}
\centering
\caption{Training hyperparameters.}
\label{table:hyperparameters_details}
\setlength{\tabcolsep}{2pt} %
\resizebox{\columnwidth}{!}{%
\begin{tabular}{lclc}
\toprule
\multicolumn{1}{l}{Parameter (PPO)} & \multicolumn{1}{c}{Value} & \multicolumn{1}{l}{Parameter} & \multicolumn{1}{c}{Value} \\ \midrule
Bias-var tradeoff ($\lambda$) & 0.95                & Torque ($T$)      &   \specialcell{stiff: [$10^1$, $10^3$]\\damp: [$10^1$, $10^3$]}   \\
Disc.\ factor ($\gamma$)        & 0.99                & Latency ($L$)     & [0, 1] sec  \\
Learning rate ($\alpha$)       & $25 \times 10^{-5}$ & Noise ($N$)       & [0, 10] \%   \\
Val.\ fun.\ coeff.\ ($c_1$)    & 0.5              & EWC reg.\ const.\ $\lambda$          & $5 \times 10^{3}$         \\
Ent.\ coeff.\ ($c_2$)             & 0.0                 & EWC buffer limit      & 100  \\
Clip range ($\epsilon$)        & 0.1                 & EWC replay batch & 32   \\
Epochs                         & 10                  & EWC replay samples    & 5000 \\
Mini-batches                   & 32                  & online-EWC $\gamma$   & 0.95 \\\toprule
\end{tabular}%
}
\end{table}

We extend the PPO~\cite{Schulman2017Proximal} implementation in \emph{StableBaselines3}~\cite{Hill2018Stable} to implement the methods described in \mysection{sec:method}.
We use the default network architecture (2-layer MLP, 64 units/layer, $tanh$ activations). Due to the different state and action space dimensions, the network for \emph{grasper} is much larger than that for \emph{reacher}.
We use the same PPO-specific hyperparameters as in~\cite{josifovski2022analysis}. For the EWC hyperparameters, we start from literature-recommended values~\cite{kirkpatrick_overcoming_2017} and adjust them empirically. All hyperparameters are listed in \mytable{table:hyperparameters_details}.

For \emph{reacher}, we use an episode length of 500 timesteps, with a timestep duration of 20ms (50Hz control frequency), for a total simulated time of 10s per episode. The maximal allowed joint speed is 20 deg/sec.
For \emph{grasper}, due to safety concerns on the real system, we reduce the maximum joint speed to 10 deg/sec, adapt the timestep duration to 50ms (20Hz control frequency), and the episode length to 200 timesteps, for the same total simulated time per episode of 10s. 
Under the \emph{ideal} training strategy, the simulator has all randomization parameters disabled throughout the entire training duration, i.e., there is no latency or noise, and the commanded joint velocities are reached instantaneously. Under the fully randomized strategy, all randomization parameters are simultaneously enabled. 
We start with a model pre-trained in ideal simulation for the CDR and \textit{finetuning} strategies. 
Then, we train them sequentially on each of the $N$, $L$, and $T$-only randomized simulations for equal duration. For the reaching task, the total training duration is $4\times 10^6$ timesteps, while for the grasping task it is $10\times 10^6$ timesteps due to the higher task complexity and a larger number of trainable parameters.  

\subsection{Evaluation Procedure and Metrics}
While the training takes place only in simulation, evaluation takes place both in simulation (ideal setting) and on the real robot, see \myfigure{fig:sim_real_envs}. 
In both cases, the same episode duration, control frequency, robot's starting configuration, maximal joint speeds, joint limits, and object/target positions per task are used, leading to identical conditions in the simulation and the real system, such that any difference in performance of the evaluated model can be attributed to the \simtoreal{} gap. 
We report multiple metrics detailed in \mytable{table:eval_metrics}, both in simulation and the real system. 

\blockcomment{
\renewcommand{\theenumi}{(\roman{enumi}}%
\begin{enumerate}
  \item \textbf{Cumulative episodic reward} ($r_\mathrm{episode}$): This is simply the sum of the rewards collected by the agent during an evaluation rollout of T steps.
  \begin{equation}
      r_\mathrm{ep} = \sum\limits_{t=0}^{T} r_t
  \end{equation}
  \item \textbf{Continuity cost} ($\mathcal{C}$)~\cite{raffin2022ssmooth}: This metric measures the smoothness of the robot's motion. It is defined as
  \begin{equation}
    \mathcal{C} = 100\times \mathbb{E}_t \left[ \left( \cfrac{a_{t+1}-a_t}{\Delta^a_\mathrm{max}} \right)^2\right]    
  \end{equation}
  where $a_{t+1}$ and $a_{t}$ are two consecutive actions and $\Delta^a_\mathrm{max}$ denotes the maximum difference between consecutive actions during the evaluation rollout. The values for $\mathcal{C}$ range between 0 (constant actions) to 100 (each consecutive action reaches opposing limits in the action space). Typically, lower values denote a smoother robot motion and are desirable on the real system.
  
  \item \todo{verify} \textbf{Distance to the target} ($d_\mathrm{tgt}$): We measure the average Euclidean distance of the center of the robot's end-effector $p_T$ to the target it has to reach $p_\mathrm{tgt}$ in the second half of the episode. This is related to its stabilization strength \cite{diarel} - the ability to balance and avoid jittering motions once the target is reached. Lower values denote better performance.
  \begin{equation}
    d_\mathrm{tgt} = ||p_T - p_\mathrm{tgt}||^2
  \end{equation}
  \item \todo{verify} \textbf{Grasping success} ($\mathcal{G}$): This metric measures how many times the object was successfully grasped during multiple evaluation rollouts. Values range from 0 to 1 (every grasp attempt was successful).
\end{enumerate}
}

\begin{table}[ht]
\vspace{4px}
   \setlength{\belowdisplayskip}{5pt}
    \centering
    \caption{Evaluation metrics.
    }
    \label{table:eval_metrics}
    \begin{tabular}{p{0.95\linewidth}}
      \toprule
      \textbf{Cumulative episodic reward} ($r_\mathrm{ep}$): The sum of the rewards collected by the agent during an evaluation rollout of $T$ steps.   
      \begin{equation}
        r_\mathrm{ep} = \sum\limits_{t=0}^{T} r_t 
      \end{equation}
      \\ \midrule
      \textbf{Continuity cost} ($\mathcal{C}$)~\cite{raffin2022ssmooth}: A measure of the smoothness of motion. 
      \begin{equation}
        \mathcal{C} = 100\times \mathbb{E}_t \left[ \left( \cfrac{a_{t+1}-a_t}{\Delta^a_\mathrm{max}} \right)^2\right]    
      \end{equation}
      where $a_{t+1}$ and $a_{t}$ are consecutive actions, and $\Delta^a_\mathrm{max}$ denotes the maximum difference between consecutive actions during the evaluation rollout. $\mathcal{C}$ ranges between 0 (constant actions) and 100 (each consecutive action reaches opposing limits in the action space). Lower values denote a smoother robot motion and are desirable in the real system. \\ \midrule
     \textbf{Distance to the target} ($d_\mathrm{tgt}$): The average Euclidean distance of the center of the robot's end-effector $p_t$ to the target $p_\mathrm{tgt}$ in the second half of the episode, when it is expected that the target is already reached. This is related to its stabilization strength~\cite{malmir2023diarel}, i.e., the ability to balance and avoid jittering motions around the target once it is reached. Lower values denote better performance. 
     \begin{equation}
        d_\mathrm{tgt} = \cfrac{2}{T} \sum\limits_{t=T/2}^{t=T} ||p_t - p_\mathrm{tgt}||^2
     \end{equation}
     \\ \midrule
    \textbf{Grasping success} ($\mathcal{G}$): Measures the percentage of successful grasps during multiple evaluation rollouts. Values range from 0 to 1, where one means that every grasp attempt was successful.\\
    \bottomrule
    \end{tabular}
    \vspace{-14px}
\end{table}

\section{Experiment Results}

\subsection{Reacher}\label{sec:res_reacher}

\begin{figure}[t]
\vspace{3px}
	\centering
	\includegraphics[width=0.49\textwidth]{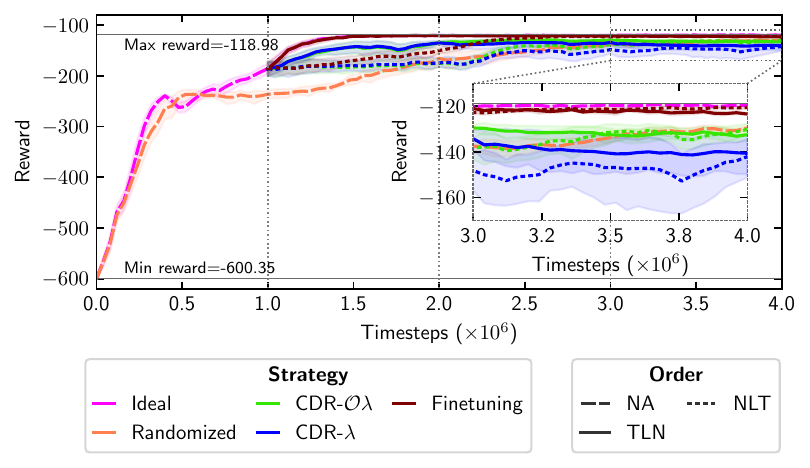}
 \vspace{-25px}
	\caption{Training progress for the reaching task. \textit{Finetuing}, \cdrewc{} and \cdronewc{} start from the \textit{Ideal} model at $10^6$ timesteps, and are sequentially trained on each randomization for $10^6$ steps, shown by vertical dotted lines. The max reward is the best reward achieved by any agent, and the min reward is the one achieved by an agent that executes random actions.}
    \label{fig:experiment_reaching_progress}
\end{figure}

The training progress of agents trained under different strategies and randomization orders to solve the \emph{reaching} task 
is presented in \myfigure{fig:experiment_reaching_progress}. Each model is trained ten times with independent seeds and their performance throughout training is evaluated in non-randomized simulation. The \emph{reaching} task is relatively easy to solve, especially in the absence of randomizations, which is why it takes less time for the \emph{Ideal} baseline to converge w.r.t.\ the \emph{Ramdomized} one.
For the sequentially trained models 
the ordering of the randomization parameters greatly affects the extent to which the policy can be improved. 
The policy improves faster when torque randomization is first ($TLN$) due to the lower uncertainty in the agent's state compared to the sequence starting with noise randomization ($NLT$). 
\emph{Finetuning} adapts much faster than the CL models (\emph{\cdrewc{}, \cdronewc{}}) and reaches performance in simulation similar to the \emph{Ideal} model. 
Among the CL models, \emph{\cdrewc{}} changes more slowly and reaches a lower performance in simulation compared to \emph{\cdronewc{}} since there is a separate regularization for each randomization parameter under EWC training, which reduces the ability of the policy to change. \emph{\cdronewc{}} adapts faster and converges to a similar performance as the fully-randomized models.   

The evaluation results in a non-randomized simulation and on the real system for each randomization strategy for the \emph{reaching} task at the end of training are presented in \mytable{table:experimentResultsReacher}. 
The \emph{Ideal} model performs best in simulation. However, it overfits to the simulation, and this leads to poor transfer to the real system, resulting in lower rewards ($r_\mathrm{ep}$), high continuity cost ($\mathcal{C}$), and unstable jittering motions around the target which results in high average distance to the target ($d_\mathrm{tgt}$). 
\emph{Randomized} has lower rewards in simulation but achieves very good \simtoreal{} transfer, leading to higher rewards, low continuity cost, and lowest distance to target on the real system. 
The CDR models also achieve very good \simtoreal{} transfer and perform comparably to or better than the fully \emph{Randomized} baseline. Due to the stronger EWC regularization during training, \cdrewc{} reaches lower rewards and higher distance to target w.r.t.\ the \cdronewc{} strategy, while it has comparable continuity cost to \cdronewc{}.
Overall, \cdronewc{} achieves the best rewards on the real system and has distance and continuity cost values similar to the fully \emph{Randomized} baseline. \emph{Finetuning} performs comparably to \emph{Ideal} in simulation but better on the real system. However, due to the lack of any regularization, it finds over-confident policies in simulation, leading to much worse \simtoreal{} transfer than the CDR strategies. The qualitative differences in performance between the different strategies are also noticeable in the supplementary video.

\blockcomment{
    \setlength{\tabcolsep}{1.5mm}
    \begin{table*}[t]
        \centering
        \caption{Reacher Results.}
        \label{table:experimentResultsReacher}
        \begin{tabular}{p{29mm} ccccccc@{}}
        \hline\noalign{\smallskip}
         \specialcell{\\Strategy}  &  \specialcell{Reward\\Simulation}    &  \specialcell{Reward\\Real} & \specialcell{Distance\\Simulation} & \specialcell{Distance\\Real} & \specialcell{Continuity Cost\\Simulation} & \specialcell{Continuity Cost\\Real}\\
        \hline\noalign{\smallskip}
            continualNLT  & -140.588  $\pm$  15.207  & -157.242  $\pm$  17.235  &  0.064  $\pm$  0.060  &  0.069  $\pm$  0.063  &  \textbf{0.000  $\pm$  0.000 } &  \textbf{0.001  $\pm$  0.000}  \\
            continualTLN  & -140.366  $\pm$  15.152  & -157.720  $\pm$  18.589  &  0.059  $\pm$  0.050  &  0.066  $\pm$  0.055  &  0.001  $\pm$  0.000  &  0.001  $\pm$  0.001  \\
     continualOnlineNLT  & -128.822  $\pm$   4.883  & \textbf{-143.881  $\pm$   4.817}  &  0.019  $\pm$  0.014  &  0.021  $\pm$  0.016  &  0.001  $\pm$  0.000  &  0.002  $\pm$  0.001  \\
     continuaOnlineTLN  & -130.965  $\pm$   4.265  & -146.953  $\pm$   4.175  &  0.028  $\pm$  0.012  &  0.032  $\pm$  0.014  &  0.001  $\pm$  0.000  &  0.002  $\pm$  0.001  \\
           finetuningNLT  & -119.574  $\pm$   0.286  & -197.777  $\pm$   7.234  &  0.001  $\pm$  0.001  &  0.188  $\pm$  0.013  &  0.005  $\pm$  0.001  &  0.159  $\pm$  0.035  \\
           finetuningTLN  & -121.857  $\pm$   1.091  & -196.752  $\pm$   8.244  &  0.009  $\pm$  0.004  &  0.183  $\pm$  0.015  &  0.004  $\pm$  0.001  &  0.108  $\pm$  0.026  \\
                   ideal  & \textbf{-119.265  $\pm$   0.382}  & -208.455  $\pm$   2.166  &  \textbf{0.001  $\pm$  0.000}  &  0.201  $\pm$  0.007  &  0.009  $\pm$  0.001  &  0.301  $\pm$  0.037  \\
              randomized  & -129.158  $\pm$   1.937  & -146.520  $\pm$   3.109  &  0.017  $\pm$  0.005  &  \textbf{0.020  $\pm$  0.006 } &  0.001  $\pm$  0.000  &  \textbf{0.001  $\pm$  0.000}  \\
    \hline\noalign{\smallskip}
        \end{tabular}

\end{table*}
}

\begin{table}[t]
    \centering
    \caption{Reacher evaluation in simulation and reality. Metrics are Episodic reward $r_\mathrm{ep}$, dist. to target $d_\mathrm{tgt}$, and continuity cost $\mathcal{C}$.}
    \label{table:experimentResultsReacher}
    \resizebox{\columnwidth}{!}{%
    \begin{tabular}{lcccccc}
\toprule
 & \multicolumn{2}{c}{$r_\mathrm{ep} \uparrow$} & \multicolumn{2}{c}{$d_\mathrm{tgt} \downarrow$} & \multicolumn{2}{c}{$\mathcal{C} \downarrow$} \\
Strategy & Sim & Real & Sim & Real & Sim & Real \\
\midrule
\cdrewc{} (NLT) & -140.59$\pm$15.21 & -157.24$\pm$17.23 & 0.06 & 0.07 & \bfseries 0.00 & \bfseries 0.00 \\
\cdrewc{} (TLN) & -140.37$\pm$15.15 & -157.72$\pm$18.59 & 0.06 & 0.07 & \bfseries 0.00 & \bfseries 0.00 \\
\cdronewc{} (NLT) & -128.82$\pm$4.88 & \bfseries -143.88$\pm$4.82 & 0.02 & \bfseries 0.02 & \bfseries 0.00 & \bfseries 0.00 \\
\cdronewc{} (TLN) & -130.97$\pm$4.26 & -146.95$\pm$4.18 & 0.03 & 0.03 & \bfseries 0.00 & \bfseries 0.00 \\
Finetuning (NLT) & -119.57$\pm$0.29 & -197.78$\pm$7.23 & \bfseries 0.00 & 0.19 & 0.01 & 0.16 \\
Finetuning (TLN) & -121.86$\pm$1.09 & -196.75$\pm$8.24 & 0.01 & 0.18 & \bfseries 0.00 & 0.11 \\
Ideal & \bfseries -119.27$\pm$0.38 & -208.46$\pm$2.17 & \bfseries 0.00 & 0.20 & 0.01 & 0.30 \\
Randomized & -129.16$\pm$1.94 & -146.52$\pm$3.11 & 0.02 & \bfseries 0.02 & \bfseries 0.00 & \bfseries 0.00 \\
\bottomrule
\end{tabular}

    }
    \vspace{-10px}
\end{table}

\blockcomment{
    \begin{figure}[htbp]
    	\centering
    	\includegraphics[width=0.5\textwidth]{figures/reacher_task2-fisher_scatter.pdf}
    	\caption{Fisher analysis for reaching task \TODO{show in comparison to finetuning}}
        \label{fig:experiment_reaching_fisher}
    \end{figure}
}

\blockcomment{
    \begin{table*}[t]
        \centering
        \caption{Grasper Results EL 5000(tbc).}
        \label{table:experimentResultsGrasperEL5000}
        \begin{tabular}{p{29mm} ccccccc@{}}
        \hline\noalign{\smallskip}
         \specialcell{\\Strategy}  &  \specialcell{Reward\\Simulation}    &  \specialcell{Reward\\Real} & \specialcell{Cont. Cost\\Simulation} & \specialcell{Cont. Cost\\Real} & \specialcell{Success\\Simulation} & \specialcell{Success\\Real}\\
        \hline\noalign{\smallskip}
            continualNLT  & -25.842  $\pm$  15.096  & -24.269  $\pm$   7.961  &  0.056  $\pm$  0.020  &  0.155  $\pm$  0.317  &  0.400  &  0.2  \\
            continualTLN  & -27.633  $\pm$  20.247  & -23.248  $\pm$   6.892  &  0.078  $\pm$  0.031  &  0.108  $\pm$  0.218  &  0.400  &  0.2  \\
     continualOnlineNLT  & -12.736  $\pm$   6.336  & -22.776  $\pm$  21.966  &  0.053  $\pm$  0.033  &  0.122  $\pm$  0.246  &  \textbf{0.784}  &  \textbf{0.6}  \\
     continualOnlineTLN  & -44.806  $\pm$  32.465  & -23.606  $\pm$   6.112  &  0.079  $\pm$  0.023  &  0.087  $\pm$  0.136  &  0.228  &  0.2  \\
           finetuningNLT  & -12.956  $\pm$   5.993  & \textbf{-14.991  $\pm$   5.558}  &  0.070  $\pm$  0.031  &  0.085  $\pm$  0.120  &  0.592  &  0.4  \\
           finetuningTLN  & -42.883  $\pm$  35.367  & -27.235  $\pm$  15.296  &  0.081  $\pm$  0.020  &  0.022  $\pm$  0.007  &  0.328  &  0.4  \\
                   ideal  &\textbf{ -12.425  $\pm$   7.689}  & -32.899  $\pm$  41.402  &  0.056  $\pm$  0.031  &  0.453  $\pm$  0.282  &  0.772  &  \textbf{0.6}  \\
              randomized  & -28.416  $\pm$   6.000  & -25.893  $\pm$   7.276  &  \textbf{0.044  $\pm$  0.040}  &  \textbf{0.005  $\pm$  0.001}  &  0.200  &  0.4  \\
    \hline\noalign{\smallskip}
        \end{tabular}
    \end{table*}
}

\begin{figure}[t]
\vspace{3px}
	\centering
	\includegraphics[width=0.49\textwidth]{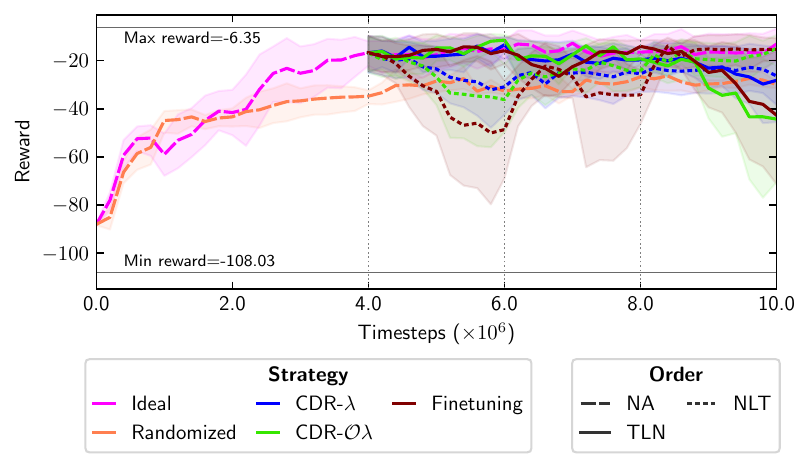}
  \vspace{-25px}
	\caption{Training progress for the grasping task. \textit{Finetuning}, \cdrewc{} and \cdronewc{} start from the \textit{Ideal} model at $4\times10^6$ timesteps, and are sequentially trained on each randomization for $2\times10^6$ steps (vertical dotted lines). The max reward is the best reward achieved by any agent, and the min reward is the one achieved by an agent that executes random actions.}
    \label{fig:experiment_grasaping_progress}
\end{figure}

\begin{table}[t]
    \centering
    \caption{Grasper evaluation in simulation and reality. Metrics are Episodic reward $r_\mathrm{ep}$, continuity cost $\mathcal{C}$, and grasping success $\mathcal{G}$.}
    \label{table:experimentResultsGrasperEL5000}
    \resizebox{\columnwidth}{!}{%
    \begin{tabular}{lcccccc}
\toprule
 & \multicolumn{2}{c}{$r_\mathrm{ep} \uparrow$} & \multicolumn{2}{c}{$\mathcal{C} \downarrow$} & \multicolumn{2}{c}{$\mathcal{G} \uparrow$} \\
Strategy & Sim & Real & Sim & Real & Sim & Real \\
\midrule
\cdrewc{} (NLT) & -25.84$\pm$15.10 & -24.27$\pm$7.96 & 0.06 & 0.15 & 0.40 & 0.20 \\
\cdrewc{} (TLN) & -27.63$\pm$20.25 & -23.25$\pm$6.89 & 0.08 & 0.11 & 0.40 & 0.20 \\
\cdronewc{} (NLT) & -12.74$\pm$6.34 & -22.78$\pm$21.97 & 0.05 & 0.12 & \bfseries 0.78 & \bfseries 0.60 \\
\cdronewc{} (TLN) & -44.81$\pm$32.47 & -23.61$\pm$6.11 & 0.08 & 0.09 & 0.23 & 0.20 \\
Finetuning (NLT) & -12.96$\pm$5.99 & \bfseries -14.99$\pm$5.56 & 0.07 & 0.09 & 0.59 & 0.40 \\
Finetuning (TLN) & -42.88$\pm$35.37 & -27.24$\pm$15.30 & 0.08 & 0.02 & 0.33 & 0.40 \\
Ideal & \bfseries -12.43$\pm$7.69 & -32.90$\pm$41.40 & 0.06 & 0.45 & 0.77 & \bfseries 0.60 \\
Randomized & -28.42$\pm$6.00 & -25.89$\pm$7.28 & \bfseries 0.04 & \bfseries 0.01 & 0.20 & 0.40 \\
\bottomrule
\end{tabular}

    }
    \vspace{-15px}
\end{table}

\subsection{Grasper}

The results of the training progress of the different strategies for the \emph{grasping} task evaluated in the non-randomized simulation are presented in \myfigure{fig:experiment_grasaping_progress}. Each experiment is repeated five times with independent seeds.
Compared to \emph{reaching}, the \emph{grasping} task is more challenging to solve due to 
the larger model sizes, the interaction with the object, and the high precision needed for successful grasping. 

With the \emph{Ideal} model,  
grasping the object and achieving a high reward is easier, while with the fully \emph{Randomized} model, it is more difficult due to the high uncertainty of the agent's state. 
Among the sequential training strategies, it can be seen that the task is significantly more challenging to solve when the noise randomization $N$ is encountered in the $TLN$ or $NLT$ sequences. The policy found at the previous randomization step significantly changes to adapt to the noise-randomized simulation. 
Overall, \emph{Finetuning} has higher deviations during training, while the deviations are lower and more stable throughout training for the CDR strategies, with \emph{\cdrewc{}} being more consistent in comparison to \emph{\cdronewc{}}. This is likely due to the separate regularization for each randomization parameter.

The evaluation results for \emph{grasping} in ideal simulation and the real world after training in the simulation are presented in \mytable{table:experimentResultsGrasperEL5000}. 
Similar to the \emph{reaching} task, \emph{Ideal} achieves high reward in simulation but achieves low average reward and high continuity cost (unsafe jittering behavior, see supplementary video) on the real system, even though the object is grasped in most runs. 
\emph{Randomized} has moderate success in finding a good policy in simulation and has the lowest grasp success in non-randomized simulation. However, it has good \simtoreal{} transfer and achieves the lowest continuity cost. 
From the sequential strategies,  
we notice that they are sensitive to the ordering and achieve higher success rates but also higher continuity cost when torque is randomized last in the sequence ($NLT$), and lower continuity cost but also lower grasp success when noise is at the end of the sequence ($TLN$). 
If we look at the rewards on the real system, \emph{Finetuning} deviates significantly and, depending on the sequence, can achieve the best or worst reward from any model that involves randomization, with a difference of about 12\%. On the other hand, the CDR models are less susceptible to the ordering of randomization, as they achieve similar rewards regardless of the sequence. Moreover, the rewards of the CDR models lie within the range of the best and worst \emph{Finetuning} strategy and are better than the reward of the \emph{Randomized} strategy.

\subsection{Effect of Continual Learning Regularization Strength}

In the context of regularization-based CL methods used to implement CDR, the regularization strength $\lambda$ plays a key role. A very low value makes the CDR models behave like \emph{Finetuning}, while a very high value makes it difficult to adapt to new randomizations after being initially trained in ideal simulation. 
Taking this into account, in our main experiments we have used a value for $\lambda$ solely based on empirical evaluation in simulation in order to maintain the zero-shot sim2real condition. Here, we perform additional experiments to provide better intuition on a suitable range for $\lambda$.

\begin{figure}[b]
\vspace{-15px}
	\centering

	\includegraphics[width=0.5\textwidth]{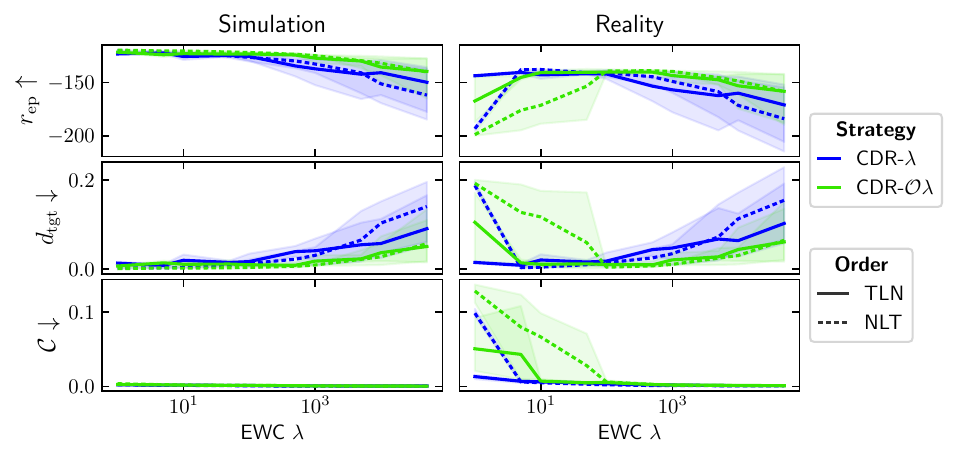}
        \vspace{-20px}

	\caption{Effect of the EWC regularization constant $\lambda$ on episodic reward~($r_\mathrm{ep}$), distance to target~($d_\mathrm{tgt}$) and continuity cost~($\mathcal{C}$) for \simtoreal{} transfer.}
    \label{fig:ablation_ewc}
\end{figure}

For this study, we employ the same training procedure as in the \emph{reaching} experiment in \mysection{sec:res_reacher}. We use the same hyperparameters (\mytable{table:hyperparameters_details}) and only change the value of the EWC $\lambda$ parameter, testing a range of different values of $\lambda = x \times 10^y$ with  $x \in \{1, 5\}$ and $y \in \{0,\cdots,4\}$. We train models of \emph{\cdrewc{}} and \emph{\cdronewc{}} in the ideal simulation followed by two different randomization sequences $TLN$ and $NLT$. We repeated them three times with independent seeds. The evaluation results of the models in the ideal simulation and on the real system are shown in \myfigure{fig:ablation_ewc}.

CDR models perform well in simulation for lower values of $\lambda$. However, they become sensitive to the order of randomizations and transfer poorly to the real system. On the other hand, higher values of $\lambda$ prevent changes to the model pre-trained on non-randomized simulation used as a starting point and perform poorly both in simulation and on the real system. In this case, the continuity cost on the real system is low because the agent performs slow moves toward the target but does not learn to balance and overshoots the target. The middle ranges of $\lambda$ perform reasonably well in simulation and give the best results on the real system while being relatively robust to the order of randomizations.

\blockcomment{
\begin{figure}[htbp]
	\centering
	\includegraphics[width=0.5\textwidth]{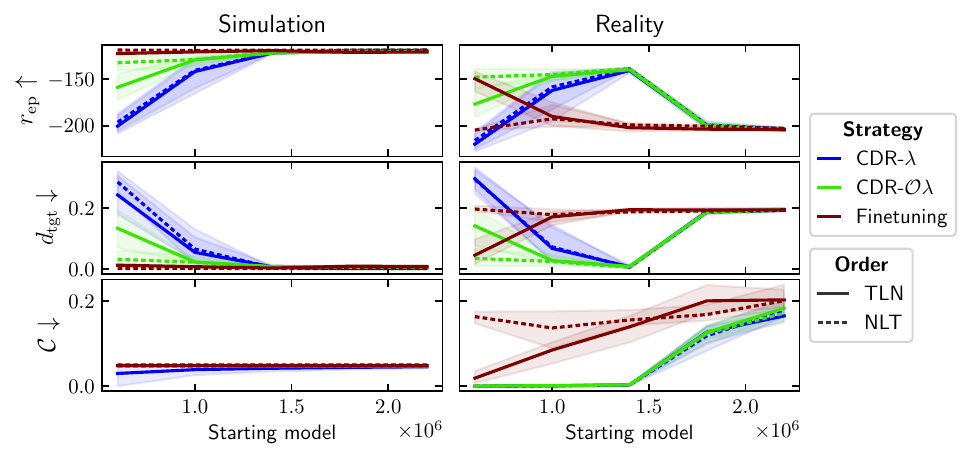}
	\caption{Effect of the ideal model's training duration on episodic reward~($r_\mathrm{ep}$), distance to target~($d_\mathrm{tgt}$) and continuity cost~($\mathcal{C}$) for \simtoreal{} transfer.}
    \label{fig:ablation_start}
\end{figure}

\subsubsection{Exploration Strategy}

\begin{figure}[htbp]
	\centering
	\includegraphics[width=0.5\textwidth]{figures/ablation_exploration_type.pdf}
	\caption{Ablation: Exploration type}
    \label{fig:ablation_explore}
\end{figure}
}

\section{Discussion and Conclusion}

In general, full randomization of many parameters simultaneously makes the task more difficult \cite{OpenAI2019Solving, Mehta2020active, josifovski2022analysis}, and our experiments show that it takes longer for the models to converge under this setting. 
Sequential randomization starting from a model pre-trained on non-randomized simulation offers a middle ground between no randomization and full randomization. However, the ordering of the randomizations can affect \simtoreal{} transfer. 
Unfortunately, in zero-shot transfer (without testing on the target system), the effect of each randomization parameter cannot be known with certainty. This can be detrimental since common sequential randomization strategies overfit to the latest randomization they were trained on. 

CDR is less susceptible to the order of randomizations and can transfer the effects of earlier randomizations in the sequence to the final model. 
This can be done either explicitly with \cdrewc{} where the transfer comes at the price of linear increase of the memory, or implicitly with \cdronewc{} where memory does not increase but the regularization effect for continual learning is weaker.
This paper uses CDR for \emph{zero-shot} \simtoreal{} transfer. However, CDR can also be adapted to provide more flexible \emph{domain adaptation} behavior by viewing the real system as the final randomization in the training sequence. EWC-$\lambda$ can be set to 0 so that the policy is finetuned to the real system dynamics or to values greater than 0 in cases where the system parameters can vary over time due to wear and tear or external conditions and the policy needs to stay more general. 

A limitation of our current training approach on one randomization at a time is that we factorize the space of randomization parameters. The complex interactions between separate randomizations in the sequence may not be captured, which may lead to sub-optimal solutions. A possible remedy for this is to define the randomization tasks as sets of different and possibly intersecting groups of randomization parameters~\cite{kadokawa2023cyclic}. Furthermore, this paper does not address how to define the ranges for the randomization parameters. Excessive randomization is possible even when a single parameter is randomized at a time, e.g., 10\% noise randomization translates to $\pm$~200mm uncertainty of the box's position and makes the grasping task challenging to solve, leading to poor grasp success on the real system. To address this issue, CDR could be combined with \emph{automated domain randomization}~\cite{OpenAI2019Solving} or \emph{active domain randomization}~\cite{Mehta2020active} to find suitable ranges for each randomization parameter. 

In summary, CDR provides a flexible framework for zero-shot \simtoreal{} transfer that does not require all randomization parameters to be defined or implemented ahead of time. It enables randomization parameter decoupling and continual model adaptation on new randomizations if required. In future work, we want to address the limitations mentioned above, as well as to implement CDR with other RL and regularization-based CL approaches and apply it in the context of vision-based domain randomization.

\bibliographystyle{IEEEtran}
\bibliography{bib_merged_cleaned.bib}

\end{document}